# Evolutionary Design of Digital Circuits Using Genetic Programming

S. M. Ashik Eftakhar, SK. Mahbub Habib, and M. M. A. Hashem
Department of Computer Science and Engineering
Khulna University of Engineering and Technology,
Khulna-9203, Bangladesh.
Ph: +880-41-774318, Fax +880-41-774403
E-mail: **mma_hashem@hotmail.com**

## Abstract

For simple digital circuits, conventional method of designing circuits can easily be applied. But for complex digital circuits, the conventional method of designing circuits is not fruitfully applicable because it is time-consuming. On the contrary, Genetic Programming is used mostly for automatic program generation. The modern approach for designing Arithmetic circuits, commonly digital circuits, is based on Graphs. This graph-based evolutionary design of arithmetic circuits is a method of optimized designing of arithmetic circuits. In this paper, a new technique for evolutionary design of digital circuits is proposed using Genetic Programming (GP) with Subtree Mutation in place of Graph-based design. The results obtained using this technique demonstrates the potential capability of genetic programming in digital circuit design with limited computer algorithms. The proposed technique, helps to simplify and speed up the process of designing digital circuits, discovers a variation in the field of digital circuit design where optimized digital circuits can be successfully and effectively designed .

**Category:** Artificial Intelligence

**Contact Person:**  Prof. Dr. M. M. A. Hashem
Head of the Department
Department of Computer Science and Engineering
Khulna University of Engineering and Technology,
Khulna-9203, Bangladesh.
 Mobile: 0171187270
Ph: +880-41-774318, Fax +880-41-774403
E-mail: **mma_hashem@hotmail.com**

**Presenter:** S. M. Ashik Eftakhar
Department of Computer Science and Engineering
Khulna University of Engineering and Technology,
Khulna-9203, Bangladesh.
E-mail: **ashik_bitk@yahoo.com.**

# Evolutionary Design of Digital Circuits Using Genetic Programming

**Abstract**—*For simple digital circuits, conventional method of designing circuits can easily be applied. But for complex digital circuits, the conventional method of designing circuits is not fruitfully applicable because it is time-consuming. On the contrary, Genetic Programming is used mostly for automatic program generation. The modern approach for designing Arithmetic circuits, commonly digital circuits, is based on Graphs. This graph-based evolutionary design of arithmetic circuits is a method of optimized designing of arithmetic circuits. In this paper, a new technique for evolutionary design of digital circuits is proposed using Genetic Programming (GP) with Subtree Mutation in place of Graph-based design. The results obtained using this technique demonstrates the potential capability of genetic programming in digital circuit design with limited computer algorithms. The proposed technique, helps to simplify and speed up the process of designing digital circuits, discovers a variation in the field of digital circuit design where optimized digital circuits can be successfully and effectively designed .*

## 1. INTRODUCTION

The research in Genetic Programming has been vastly developed over the last few years. Genetic Programming has recently been applied for designing analog electrical circuits. In the work of Koza et al., on analog circuit synthesis by means of genetic programming (GP), the component values, number of components and the circuits topologies are evolved together, where various analog filter design and circuit synthesis problems have been solved using this approach [1]. Like analog circuits, the design of digital circuits is not really an easy task. These digital circuits are inseparable part of the modern computer life. What is done with the computer is basically computation and logic-based operations. For this reason, digital circuits are badly needed. But designing digital circuits is generally a complicated and time-consuming task requiring knowledge of large collections of domain-specific rules. The design basically depends on the understanding of operations of the circuits. To design a digital circuit efficiently and with little algorithmic knowledge, our traditional digital circuit solution strategy must be changed. For this reason, like the automatic synthesis of analog electrical circuits used in [1], [2], we are trying to design digital circuits with genetic programming efficiently and effectively.

The conventional digital circuit design approach involves a large amount of task to develop digital circuits, whereas genetic programming can design a digital circuit heuristically without performing a large number of tasks. The tasks that the conventional approach includes the statement of the problem, determination of number of input and output variables, the letter symbols for input and output variables, the truth table defining input-output mapping, simplified Boolean expression and drawing of circuits [3]. On the other hand, the automatic synthesis just requires the definition of input and output variables, the input-output mapping and drawing of circuits. For those significant benefits of designing digital circuits using genetic programming, we have adopted this technique to design different critical digital circuits. Our objective here is to design digital circuits automatically using Genetic Programming with Evolutionary Tree Mutation. Our goal is to achieve 100% accurate result in designing digital circuits, that is, to achieve a fitness that is equal to required fitness. Here it is being tried to focus on the design of comparatively critical digital circuits used in many spheres of digital operations.

In this paper, the method of designing digital circuits using a modern approach in genetic programming with Evolutionary Tree Mutation [4] has been proposed. Here the effort, outcome and measurement of the accuracy of our outcome are represented.

## 2. GENETIC PROGRAMMING: AN EVOLUTIONARY DESIGN OVERVIEW

Finding a potential solution of the problem at hand represented by a suitable chromosome is a key issue for evolutionary computation, e.g., genetic programming. Currently, there are numerous approaches used as the chromosome representation of hardware structure [1], [6]-[11]. In [6], for example, an approach to the evolutionary design is described. The reported design method, however, is based on direct evolution with gate level primitive components such as logic gates and flip-flops. In our research, it is tried to solve the problem of designing digital circuits using Genetic Programming; hence it is needed to formulate the problem genetically. In order to formulate the problem, the basics of Genetic Programming are required. Using this technique, a new solution is searched in non-linear time, with relative certainty that the final outcome will converge on a near-optimal, global solution. Much of the theory behind genetic programming is the same as that behind genetic algorithms. The same Darwinian concept of survival-of-the-fittest applies, through genetic operators, but with a twist. Sometimes Genetic programming can solve problem more efficiently than Genetic Algorithm.

Genetic programming is proposed as a search and optimization technique for automatic program induction: a population of trial computer programs is maintained, mutation operations produce changes in these programs, and selection is used to determine which programs survives to the next generation and which programs are culled from the pool of trials. The process is repeated until an acceptable program is obtained or the allotted computer time is exhausted. The genetic programming procedure consists of an initialization followed by an iterative loop of mutation and selection.

*2.1 Initialization*

The initial population of trial computer is randomly generated. There are several methods for generating parse trees that can be used to initialize the population. Here the initialization depends upon the maximum allowed length of tree. Once the trees are generated, the performance of each of randomly generated trees is evaluated based on objective function. In our problem, we have used circuits as individuals. In the initial population, there are some defined number of circuits for us for forward propagation. Initially, what we do is to assign fitness to each individual for the further generation processing.

*2.2 Offspring Generation by Mutation*

Six different mutation operators are employed to generate offspring from parents. These mutation operators are named O*neNode, AllNodes, Swap, Grow, and Trunc* [4]. The tree structures consists of nodes, with the function nodes taking one or more arguments and terminal nodes taking zero arguments.

*2.3 Parent Selection*

*EP-style tournament selection* [4] with ten opponents is applied to select the parents for the next generation. Every program in the population was compared with ten randomly selected opponents. For each comparison in which the program receives a better than or equal score, it receives a win. The better half of the population with the largest number of wins becomes the parents for the next generation. The success predicate of the retained parents is computed to check for the discovery of a solution program that solves the problem. If the solution is found, the search procedure is terminated and the problem is considered to be solved; otherwise the process of *mutation* and *selection* is continued for the next generation.

# 3. CIRCUIT DESIGN METHODOLOGY

In this paper, said genetic programming has been used to design digital circuits using the evolutionary tree mutation technique. Subtree mutation transforms the structure of a circuit preserving its correctness property, that is, the input output mapping must be perfect. Selection is used to determine which programs survive to the next generation and which programs are culled from the pool of trials. The process is repeated until an acceptable program is obtained or the allotted computer time is exhausted. The initialization is followed by an iterative loop of mutation and selection. For our problem, the individuals are generated in the initial population as follows. First, we select functional nodes or functions randomly from the table of functions, then count the number of input and output terminals in all nodes selected, next add specific functional nodes so as to balance the input and output terminals, and finally connect the terminals randomly to generate a correct digital circuit. After the generation of initial population, the subtrees of each individual are also randomly generated [5].

*3.1 Functions and Terminals*

Functional representation includes the internal structure of a circuit. Generally, the circuits are designed internally as *parse trees*. To design a parse tree the definition of operators is a must; the operators that are used here are termed as functional nodes or functions [**TABLE I**]. Different problem uses different functions and terminals. For digital circuits, functions are *Basic gates (AND-gate, OR-gate, NOT-gate), Combinational functional elements (Half Adder-HA, Full Adder-FA etc.) and Sequential Memory elements (JK-Flip-Flop, RS-Flip-Flop, D- Flip-Flop, T- Flip-Flop etc.).* The terminals are the input variables used in the circuits. Suppose, a circuit consists of input set {*a,b*}, then *a* and *b* are *terminals*. These terminals, as well as, functions are initialized to use in the design of circuits.

TABLE I
FUNCTIONS

| NAME | FUNCTIONAL DESCRIPTION | INPUT-OUTPUT |
|---|---|---|
| OR-GATE | F = A + B | 2-Input-1-output |
| AND-GATE | F = A . B | 2-Input-1-output |
| NOT-GATE | F = ¬ A | 1-Input-1-output |
| NAND-GATE | F = ¬ (A . B) | 2-Input-1-output |
| NOR-GATE | F = ¬ (A + B) | 2-Input-1-output |
| HA (Half Adder) | F = A plus B | 2-Input-1-output |
| FA (Full Adder) | F = A plus B plus carry | 3-Input-1-output |
| JKFF(JK Flip-flop) | Memory Element | 2-Input-1-output |
| RSFF(RS Flip-flop) | Memory Element | 2-Input-1-output |
| T-FF(T Flip-flop) | Memory Element | 1-Input-1-output |
| D-FF(D Flip-flop) | Memory Element | 1-Input-1-output |

### 3.2 Fitness Calculation

The fitness calculation is mainly depends on the truth table that indicates the input-output mapping. The fitness for the circuits, generated in each generation, is compared to the desired fitness that indicates the correctness of our design. So fitness is termed as the performance measure of our design. Depending on the fitness achieved, decision can be made whether there is a variation from the possible circuit representation or how much far we are from our actual design.

### 3.3 Circuit Verification

After getting the desired fitness, a new *circuit verification algorithm* is applied to check whether the circuit is perfect or not. If the circuit is not perfect, the process of designing circuits is repeatedly applied.

---

1. Calculate total no. of combinations: $2^{inputs}$
2. For each combination -
    *i) Replace each terminal with the corresponding input terminal value by matching with input terminals.*
    *ii) Get the desired output from truth table.*
    *iii) Calculate desired output until the measured circuit is solved for each Combination-*
        a) *Calculate the function values as defined in functions.*
        b) *Update after calculating values.*
3. desired output = measured output for all combinations : Correct Circuit.
4. desired output ≠ measured output for any combination : Wrong circuit.

---

**Fig 3.1:** *Circuit verification algorithm*

# 4. IMPLEMENTATION

To implement the program of designing digital circuits in computer, there is an essence of control parameters listed in [**TABLE II**]. Besides this, there must be the alphabetic terminal names, number of inputs and outputs, and a truth table that is externally given. Moreover, randomness is performed based on time-based seed. There is also a limit for number 'runs' or 'trials'. Using these information of a circuit, a perfect digital circuit can be designed.

TABLE II
MAIN CONTROL PARAMETER VALUES

| Population size | 1000 | Weight of Allnode mutation | 100 |
|---|---|---|---|
| Gross number of generations | 1000 | Weight of Swap mutation | 100 |
| Max. number of nodes | 50 | Weight of Grow mutation | 100 |
| Initial Population depth | 10 | Weight of Trunc mutation | 100 |
| Tournament Size | 10 | Probability of mutation | 1.0 |

TABLE III
SAMPLE TRUTH TABLE FOR
3-INPUT 1-OUTPUT CIRCUIT

| Inputs | | | Outputs |
|---|---|---|---|
| A2 | A1 | A0 | F |
| 0 | 0 | 0 | 0 |
| 0 | 0 | 1 | 0 |
| 0 | 1 | 0 | 0 |
| 0 | 1 | 1 | 1 |
| 1 | 0 | 0 | 1 |
| 1 | 0 | 1 | 1 |
| 1 | 1 | 0 | 1 |
| 1 | 1 | 1 | 1 |

# 5. RESULTS

The genetic programming approach has been used here to design digital circuits efficiently having small to large number of inputs. The circuits (both *combinational* and *sequential* circuits) that are designed have inputs from three to ten and outputs having one to more; circuits having more than ten inputs can also be designed properly. The circuits obtained depending on the truth table are expressed in the program as parse tree representations and they are in prefix notation. The results and the circuit diagrams for multi-input circuits are listed in **TABLE IV** and in fig 5.1-5.9.

## TABLE IV
### RESULTS OBTAINED IN GENETIC PROGRAMMING

| No. of inputs | No. of outputs | Input Terminals | Output Terminals | Functions | No. of trials | No. of Generations |
|---|---|---|---|---|---|---|
| 3 | 1 | {A0, A1, A2} | {F} | {AND, OR, NOT, HA, FA} | 10 | 1 |
| 4 | 1 | {A0, A1, A2, A3} | {F} | {AND, OR, NOT, HA, FA} | 15 | 1 |
| 4 | 2 | {A0, A1, A2, A3} | {F1,F2} | {AND, OR, NOT, HA, FA} | 10(F1) 50(F2) | 1(F1) 25(F2) |
| 5 | 1 | {A0, A1, A2, A3, A4} | {F} | {AND, OR, NOT, HA, FA, JKFF, RSFF, DFF, TFF} | 40 | 1 |
| 6 | 1 | {A0, A1, A2, A3, A4, A5} | {F} | {AND, OR, NOT, HA, FA} | 50 | 23 |
| 7 | 1 | {A0, A1, A2, A3, A4, A5, A6} | {F} | {AND, OR, NOT, HA, FA} | 30 | 60 |
| 8 | 1 | {A0, A1, A2, A3, A4, A5, A6, A7} | {F} | {AND, OR, NOT, HA, FA} | 30 | 6 |
| 9 | 1 | {A0, A1, A2, A3, A4, A5, A6, A7, A8} | {F} | {AND, OR, NOT, HA, FA} | 65 | 40 |
| 10 | 1 | {A0, A1, A2, A3, A4, A5, A6, A7, A8, A9} | {F} | {AND, OR, NOT, HA, FA} | 60 | 10 |

## A. 3-Input-1-Output Combinational Circuit

According to the sample truth table [**TABLE III**], the correct expression achieved in genetic programming is: **F** = *(AND (OR A0 A2) (OR A2 A1))*
The actual Boolean expression for the above expression found in genetic programming is: **F** = *(A0+A2).(A1+A2)*

Corresponding circuit diagram is given below.

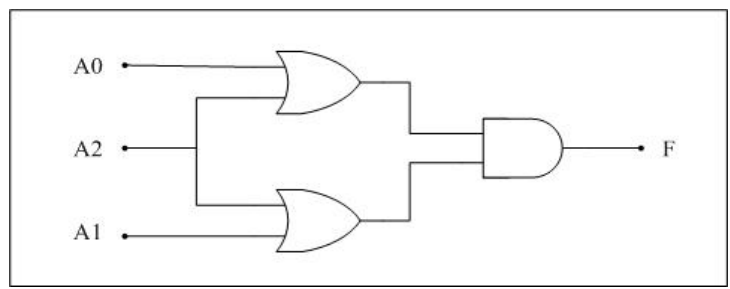

**Figure 5.1:** Circuit Diagram discovered by GP for 3-input-1-output combinational circuit

## B. *4-Input-1-Output Combinational Circuit*

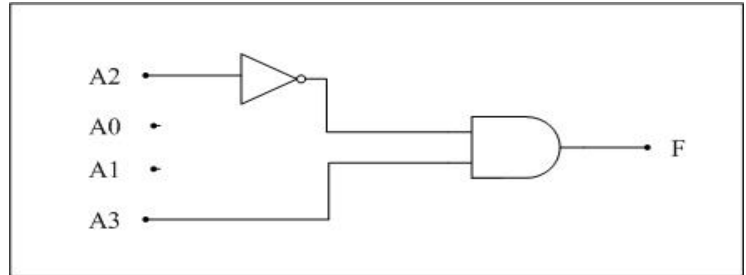

**Figure 5.2:** Circuit Diagram discovered by GP for 4-input-1-output combinational circuit for a sample truth table

## C. *4-Input-2-Output Combinational Circuit*

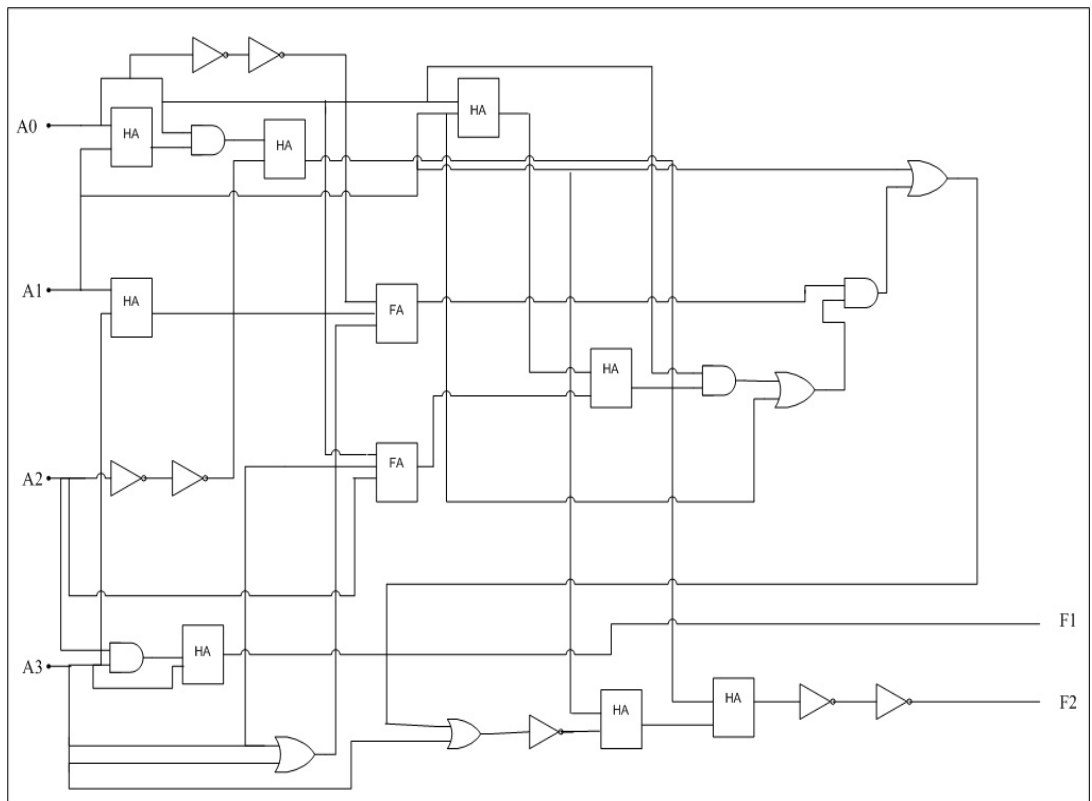

**Figure 5.3:** Circuit Diagram discovered by GP for 4-input-2-output combinational circuit for a sample truth table

## D. 5-Input-1-Output Sequential Circuit

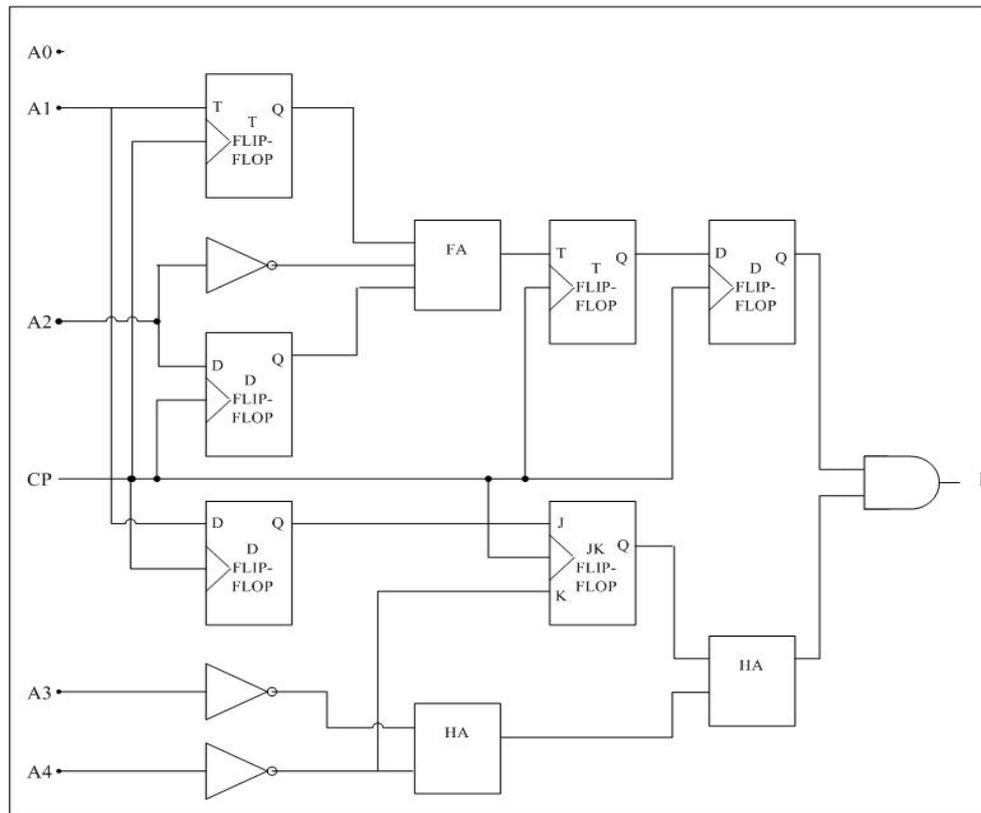

**Figure 5.4:** Circuit Diagram discovered by GP for 5-input-1-output sequential circuit for a sample truth table

## E. 6-Input-1-Output Combinational Circuit

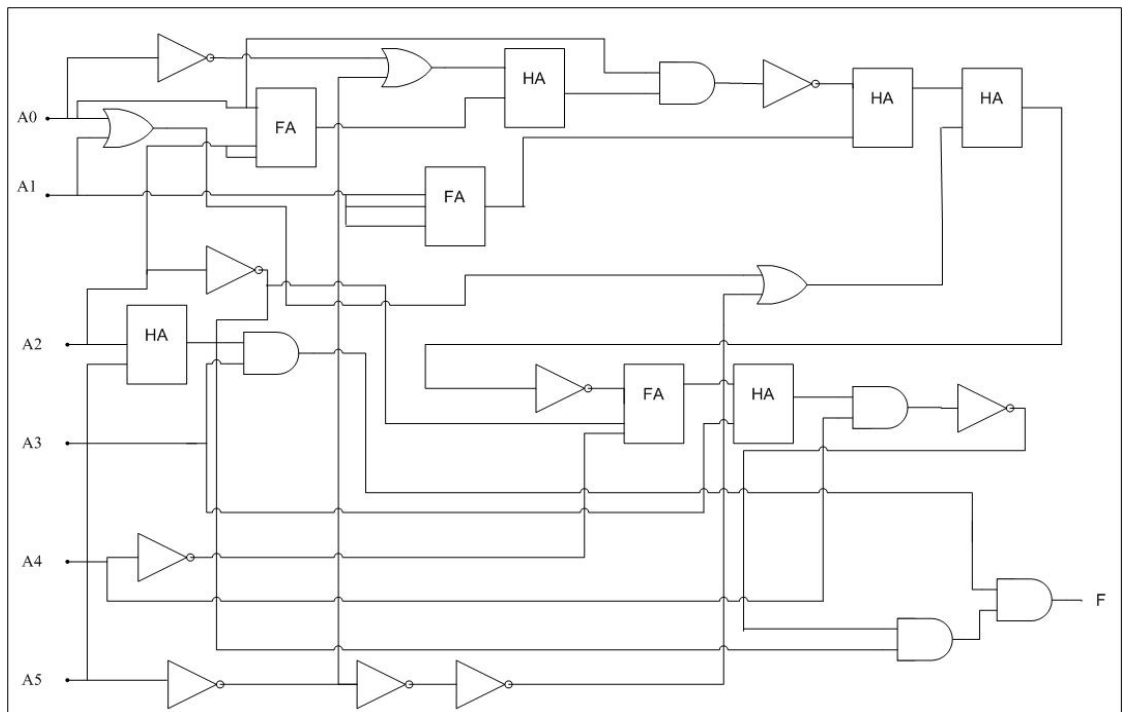

**Figure 5.5:** Circuit Diagram discovered by GP for 6-input-1-output combinational circuit for a sample truth table

## F. 7-Input-1-Output Combinational Circuit

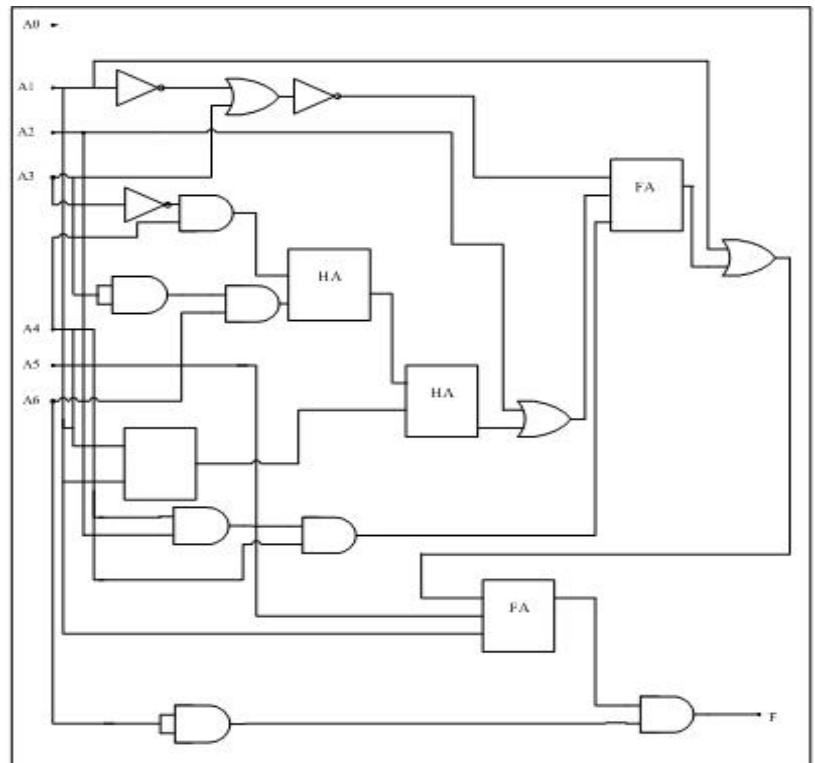

**Figure 5.6:** Circuit Diagram discovered by GP for 7-input-1-output combinational circuit for a sample truth table

## G. 8-Input-1-Output Combinational Circuit

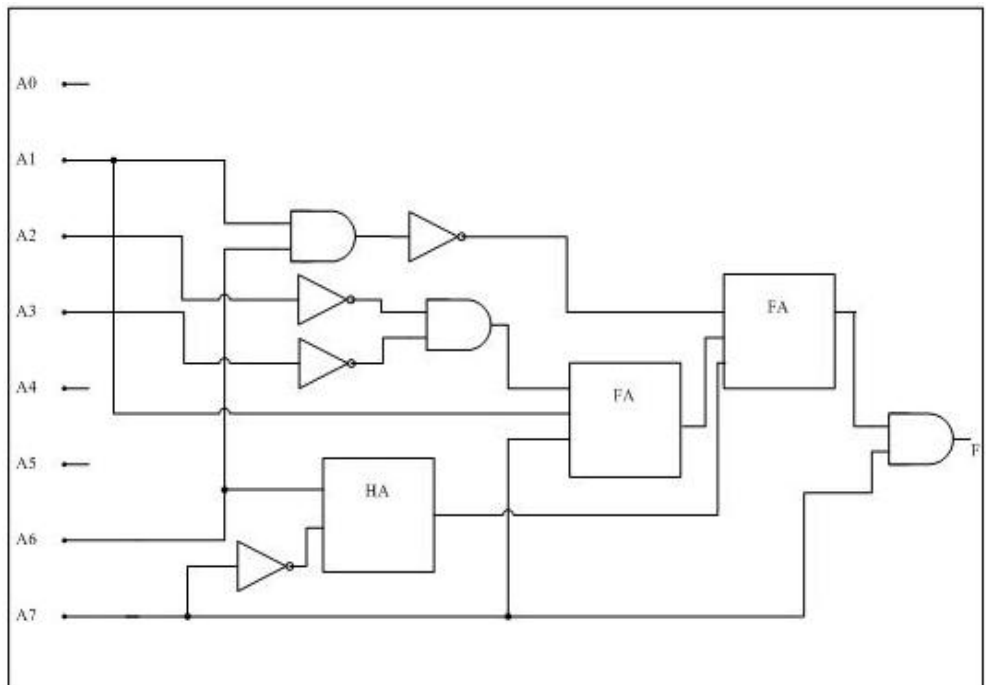

**Figure 5.7:** Circuit Diagram discovered by GP for 8-input-1-output combinational circuit for a sample truth table

## H. 9-Input-1-Output Combinational Circuit

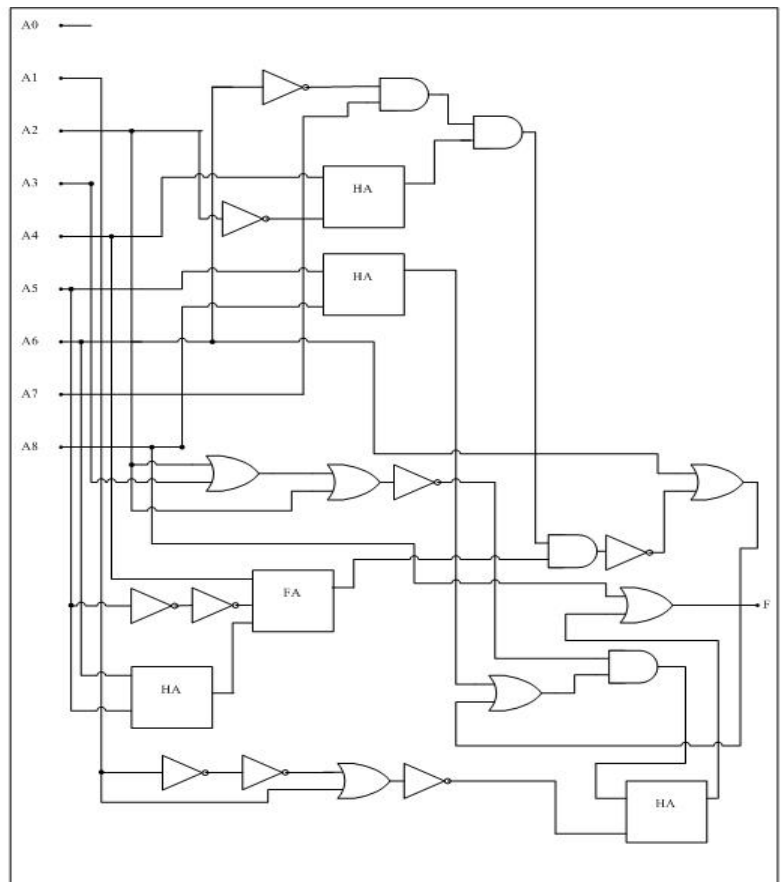

**Figure 5.8:** Circuit Diagram discovered by GP for 9-input-1-output combinational circuit for a sample truth table

## I. 10-Input-1-Output Combinational Circuit

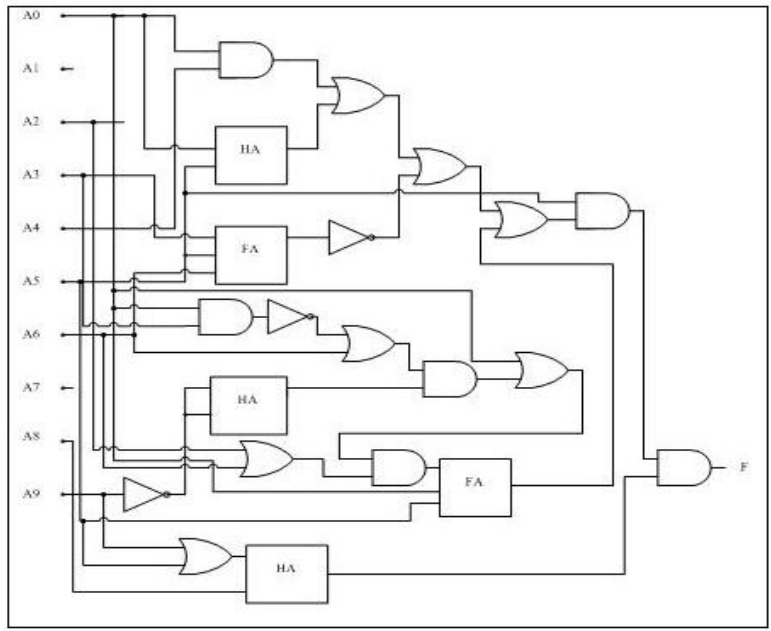

**Figure 5.9:** Circuit Diagram discovered by GP for 9-input-1-output combinational circuit for a sample truth table

Genetic programming, while designing the digital circuits, shows appreciable performance. In this approach, the fitness error decreases generation by generation. The divergence of error generation by generation is shown in fig 5.10, fig 5.11, fig 5.12, and fig. 5.13. The fitness error shown in the figure is the error with circuits that were tested and as error goes down to zero, it can be said that the circuit is correct, otherwise wrong.

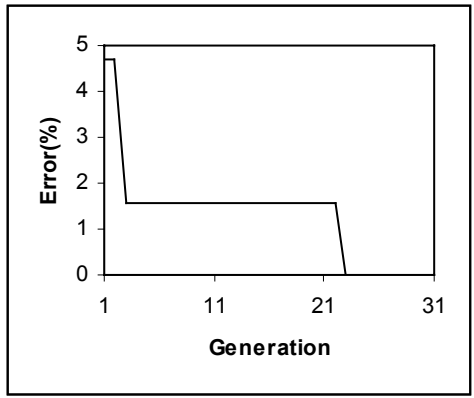

**Fig 5.10:** Divergence in error for GP for 6-input circuit showing Generation vs Error(%)

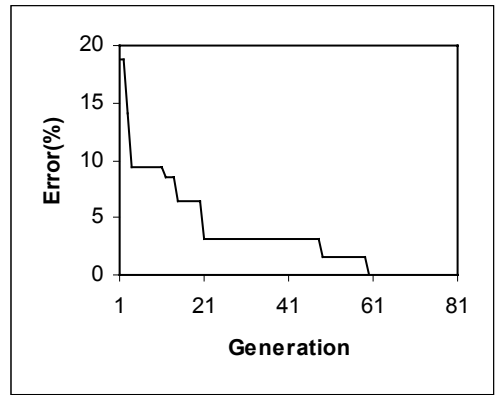

**Fig 5.11:** Divergence in error for GP for 7-input circuit showing Generation vs Error(%)

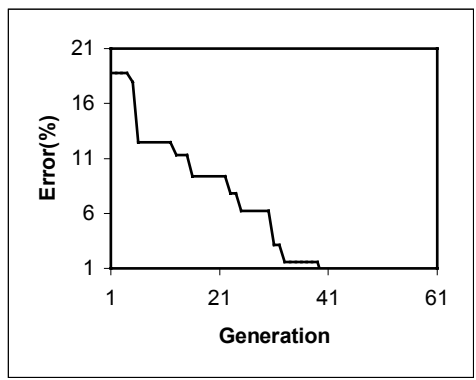

**Fig 5.12:** Divergence in error for GP for 9-input circuit showing Generation vs Error(%)

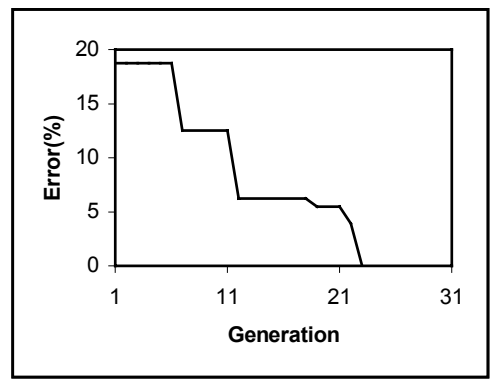

**Fig 5.13** Divergence in error for GP for 10-input circuit showing Generation vs Error(%)

## 6. CONCLUSION

In this paper, an efficient digital circuit design technique using genetic programming with subtree mutation is proposed which is applied to various combinational and sequential circuit design. In fact, the proposed technique can easily be applied to the different digital circuit design specification by changing fitness functions. In addition, the effectiveness of our design has been shown in the result. In the sample design of digital circuits, the circuit got is also verified and found that the design is perfect. This implies that the proposed design can help to speed up the process of designing digital circuits. So, Genetic programming can be a nice choice for designing more complex arithmetic circuits, like ALU design. Our proposed technique can do better in this respect.